\documentclass[journal]{IEEEtran}
\usepackage{color,soul}
\sethlcolor{lightgray}

\usepackage[none]{hyphenat} % Stop breaking up words in a table
\usepackage{array}

\newcolumntype{^}{>{\currentrowstyle}}

\usepackage{booktabs}
\usepackage{multirow}
\usepackage{threeparttable}

\usepackage{graphicx} % Allows to import images
\usepackage{float} % Allows for control of float positions
\usepackage{subfig}
\graphicspath{{figs/}}
\usepackage{tikz}
\usetikzlibrary{shapes.geometric,arrows,matrix,positioning}
\usepackage{pgfplots}

\pgfplotsset{compat=newest} 
\newcounter{plotno}

\usetikzlibrary{pgfplots.groupplots,backgrounds}
\usepgfplotslibrary{units}
\usepackage{tikzscale}

\tikzstyle{io} = [rectangle, minimum width=2cm, minimum height=2cm, text width=2cm, text centered]
\tikzstyle{block} = [rectangle, minimum width=2cm, minimum height=2cm, text width=2cm, text centered, draw=black, line width=0.3mm]
\tikzstyle{arrow_fc} = [line width=0.8mm,->,>=stealth]

\tikzstyle{sum} = [circle, minimum width=0.4cm, draw=black]
\tikzstyle{node1} = [circle, minimum width=0.4cm, draw=black, fill=brown!180]
\tikzstyle{node2} = [circle, minimum width=0.3cm, draw=black, fill=brown!180]
\tikzstyle{node3} = [circle, minimum width=0.2cm, draw=black, fill=brown!180]
\tikzstyle{leaf1} = [rectangle, minimum width=0.2cm, minimum height=0.2cm, draw=black, fill=green]
\tikzstyle{leaf2} = [rectangle, minimum width=0.2cm, minimum height=0.2cm, draw=black, fill=red]
\tikzstyle{arrow} = [line width=0.4mm,-]
\tikzstyle{arrow2} = [line width=0.6mm,->, >=stealth, color=cyan]
\tikzstyle{arrow3} = [line width=0.5mm,->, >=stealth,shorten >= 0.2cm, shorten <= 0.1cm]

\usepackage{mhchem} % Allows us to write chemisty equations
\usepackage{xfrac} % Allows for slanted fractions
\usepackage[numbers,sort&compress]{natbib}

\usepackage{setspace}
\usepackage{tikz}
\usepackage[upright]{fourier}
\usepackage{tkz-kiviat,numprint} 

\usetikzlibrary{decorations.pathreplacing, arrows, fit}
\usepackage{graphicx}
\usepackage{url}
\usepackage{booktabs}
\usepackage{textcomp}
\usepackage{epstopdf}
\usepackage{hhline}
\usepackage{xmpmulti}
\usepackage{siunitx}
\usepackage{adjustbox}

\pgfdeclarelayer{background}
\pgfdeclarelayer{foreground}
\pgfsetlayers{background,main,foreground}

\begin{document}
%\pgfplotsversion
%\large
\title{A Generalised Seizure Prediction with Convolutional Neural Networks for Intracranial and Scalp Electroencephalogram Data Analysis}

\author{Nhan~Duy~Truong, Anh~Duy~Nguyen, Levin~Kuhlmann, Mohammad~Reza~Bonyadi, Jiawe~Yang, Omid~Kavehei$^{*}$ \thanks{$^{*}$\textit{Corresponding author}: kavehei@ieee.org.}}%

%\author{Nhan~Truong,~\IEEEmembership{Student~Member,~IEEE,}
%      
%        and~Omid~Kavehei,~\IEEEmembership{Member,~IEEE}% <-this % stops a space
%}%\thanks{}% <-this % stops a space

%\thanks{$^{*}$ corresponding author: kavehei@ieee.org.}% <-this % stops a space
%\thanks{Manuscript modified May 16, 2016}}


\markboth{DEC~2017}%
{Shell \MakeLowercase{\textit{et al.}}: Bare Demo of IEEEtran.cls for IEEE Journals}



\maketitle

\begin{abstract}
%\large
Seizure prediction has attracted a growing attention as one of the most challenging predictive data analysis efforts in order to improve the life of patients living with drug-resistant epilepsy and tonic seizures. Many outstanding works have been reporting great results in providing a sensible indirect (warning systems) or direct (interactive neural-stimulation) control over refractory seizures, some of which achieved high performance. However, many works put heavily handcraft feature extraction and/or carefully tailored feature engineering to each patient to achieve very high sensitivity and low false prediction rate for a particular dataset. This limits the benefit of their approaches if a different dataset is used. In this paper we apply Convolutional Neural Networks (CNNs) on different intracranial and scalp electroencephalogram (EEG) datasets and proposed a generalized retrospective and patient-specific seizure prediction method. We use Short-Time Fourier Transform (STFT) on \boldmath$30$-second EEG windows to extract information in both frequency and time domains. A standardization step is then applied on STFT components across the whole frequency range to prevent high frequencies features being influenced by those at lower frequencies. A convolutional neural network model is used for both feature extraction and classification to separate preictal segments from interictal ones. The proposed approach achieves sensitivity of \boldmath$81.4\%$, \boldmath$81.2\%$, \boldmath$82.3\%$ and false prediction rate (FPR) of \boldmath$0.06$/h, \boldmath$0.16$/h, \boldmath$0.22$/h on Freiburg Hospital intracranial EEG (iEEG) dataset, Children's Hospital of Boston-MIT scalp EEG (sEEG) dataset, and Kaggle American Epilepsy Society Seizure Prediction Challenge's dataset, respectively. Our prediction method is also statistically better than an unspecific random predictor for most of patients in all three datasets.
\end{abstract}


% \fi
\IEEEpeerreviewmaketitle

\section{Introduction}
\IEEEPARstart{A}{dvances} in data mining and machine learning in the past few decade has attracted significantly more attention to the application of these techniques in detective and predictive data analytics especially in healthcare, medical practices and biomedical engineering \cite{Ramgopal2014szloop,Gadhoumi2016,BouAssi2017szpred}. While the body of available proven knowledge lacks a convincing and comprehensive understanding of sources of epileptic seizures, some early works showed the possibility of predicting, seemingly unpredictable, seizures \cite{rogowski1981szpred,salant1998szpred}. In ref. \cite{Maiwald2004SPH}, dynamical similarity index, effective correlation dimension and increments of accumulated energy were used as feature extraction. Dynamical similarity index yielded highest performance with sensitivity $42\%$ and false prediction rate (FPR) less than $0.15$/h. Mean phase coherence and lag synchronization index of $32$-s sliding EEG windows were used as features for seizure prediction \cite{winterhalder2006szpred}. Performance of this approach was still modest at sensitivity of $60\%$ and a comparable FPR. This approach was further improved by combining bi-variate empirical mode decomposition and Hilbert-based mean phase coherence as feature extraction \cite{zheng2014szpred}. As a result, sensitivity was increased to beyond $70\%$ while FPR dropped below $0.15$/h. Another method to exploit the synchronization information was proposed by authors in \cite{Parvez2017szpred}. In that method, phase-match error of two consecutive epochs is calculated first, then applied discrete cosine transform (DCT) on the phase-match error in order to estimate energy concentration ratio. The average of energy concentration ratio across all channels was then used as global features. The authors extracted local features based on modified deviation and fluctuation functions, and  LS-SVM was used for classification which resulted in $95.4\%$ sensitivity and $0.36$/h FPR.

A machine learning approach using Support Vector Machine (SVM) with features from nine frequency bands of spectral power was introduced in \cite{parkyun2011szpred}. This method achieved a decent performance on Freiburg Hospital dataset \cite{EEGFB} with sensitivity of $98.3\%$ and FPR of $0.29$/h. A similar approach with additional features, power spectral density ratios, was proposed by \cite{Zhang2016szpred} with very high sensitivity exceeding $98\%$ and FPR less than $0.05$/h. However, this approach extremely tailored feature selection for each patient, hence, lacked of generalization. Different from the two approaches above, \cite{aarabi2014szpred} did a Bayesian inversion of power spectral density then applied a rule-based decision to perform the seizure prediction task. This approach was tested with the same Freiburg dataset with sensitivity of $87.07\%$ and FPR of $0.2$/h. The authors, in a recent work \cite{Aarabi2017szpred}, extracted six uni-variate and bi-variate features including correlation dimension, correlation entropy, noise level, Lempel-Ziv complexity, largest Lyapunov exponent, and nonlinear interdependence and achieved a comparable sensitivity of $86.7\%$ and lower FPR of $0.126$/h.

Based on an assumption that the future events depend on a number of previous events, multi-resolution $N$-gram on amplitude patterns was used as feature extraction in \cite{eftekhar2014szpred}. After optimizing feature set per patient, this method yielded a high sensitivity of $90.95\%$ and a low FPR of $0.06$/h. Recently, \cite{Sharif2017} captured dynamics of EEG by using $64$ fuzzy rules to estimate trajectory of each sliding EEG window on Poincar{\'e} plane. The features went through PCA to reduce interrelated features before classified by a SVM. This work achieved a decent performance with sensitivity of more than $91\%$ and FPR below $0.08$/h.

Other seizure prediction techniques were proposed by \cite{Gadhoumi2012szpred,li2013szpred}. In \cite{Gadhoumi2012szpred}, features estimated by wavelet energy and entropy were optimized for each patient, then a discriminant analysis was used to separate preictal segments from interictal ones. The results were promising with sensitivity of $88.9\%$ and FPR of $0.3$/h testing with intracranial EEG data from six patients from Montreal Neurological Institute dataset. \cite{li2013szpred} introduced a lightweight approach based on spike rate. This approach was able to achieve a sensitivity of $75.8\%$ with a false prediction rate of $0.09$/h.

There have been works claimed to have $100\%$ sensitivity and very low false alarm, less than $0.05$/h \cite{Zhang2016szpred}, or even zero false alarm \cite{mirowski2008comparing}. However, these works employed numerous feature engineering techniques and seizure prediction for each patient performs well only with a certain technique. For example, in \cite{mirowski2008comparing}, the authors used $6$ different feature extraction methods and three machine learning algorithms. Similarly, in \cite{Zhang2016szpred}, there were $44$ features and a set of $91$ cost-sensitive linear SVM classifiers being used to search for the optimal single features or feature combinations that performs the best for each patient. These approaches have two main drawbacks: (1) we do not know which combination of features and classifier will work for a new patient, and (2) we cannot guarantee that the optimal combination will work well with future data of the same patient.

We are seeking for an approach that can be applied for all patients with minimum feature engineering. Neural networks are known with capability to extract features from raw input data to perform a classification task. In this work, we will deploy a convolutional neural network for seizure prediction. The main contributions of this work are: (1) propose a proper method to pre-process raw EEG data into a form suitable for convolutional neural network, and (2) propose a guideline to help convolutional neural network perform well with seizure prediction task with minimum feature engineering. To prove the advantage of our approach, we will use the same pre-processing technique and convolutional neural network configuration for all patients from two different datasets: Freiburg Hospital intracranial EEG (iEEG) dataset and Children's Hospital of Boston-Massachusetts Institute of Technology (CHB-MIT) scalp EEG (sEEG) dataset.
 

\section{Proposed Method}
\subsection{Dataset}
There are three datasets being used in this work: Freiburg Hospital dataset \cite{EEGFB}, CHB-MIT dataset \cite{shoeb2009application} and Kaggle American Epilepsy Society Seizure Prediction Challenge's dataset \cite{BenjaminKaggleSzPred2014}. The Freiburg dataset consists of intracranial EEG (iEEG) recordings of $21$ patients with intractable epilepsy. Due to lack of availability of the dataset, we are only able to use data from $13$ patients. A sampling rate of $256$~Hz was used to record iEEG signals from these $13$ patients. In this dataset, there are $6$ recording channels from $6$ selected contacts where three of them are from epileptogenic regions and the other three are from the remote regions. For each patient, there are at least $50$~min preictal data and $24$~h of interictal. More details about Freiburg dataset can be found in \cite{Maiwald2004SPH}.

CHB-MIT dataset contains scalp EEG (sEEG) data of $23$ pediatric patients with $844$~h of continuous sEEG recording and $163$ seizures. Scalp EEG signals were captured using $22$ electrodes at sampling rate of $256$~Hz \cite{shoeb2009application}. We define interictal periods that are at least $4$~h away before seizure onset and after seizure ending. In this dataset, there are cases that multiple seizures occur close to each other. For the seizure prediction task, we are interested in predicting the leading seizures. Therefore, for seizures that are less than $30$~min away from the previous one, we consider them as only one seizure and use the onset of leading seizure as the onset of the combined seizure. Besides, we only consider patients with less than $10$ seizures a day for the prediction task because it is not very critical to perform the task for patients having a seizure every $2$~h on average. With the above definition and consideration, there are $13$ patients with sufficient data (at least $3$ leading seizures and $3$ interictal hours).

Kaggle seizure prediction challenge's dataset has iEEG data of $5$ canines and $2$ patients with $48$ seizures and $627.7$~hours of interictal recording \cite{BenjaminKaggleSzPred2014}. Intracranial EEG canine data were recorded from $16$ implanted electrodes with $400$~Hz sampling rate. Recorded iEEG data of the two patients were from $15$ depth electrodes (patient 1) and $24$ subdural electrodes (patient 2) at sampling rate of $5$~kHz. Preictal and interictal $10$-min segments were extracted by the challenge's organizers. Specifically, for each lead seizure, six preictal segments were extracted from $66$~min to $5$~min prior to seizure onset with $10$~s apart in time. Interictal segments were randomly selected at least one week away from any seizure.

\subsection{Pre-processing}
\begin{figure}[h]
%\hspace*{0.1in}
\subfloat[]{%
	\includegraphics[width=0.9\columnwidth]{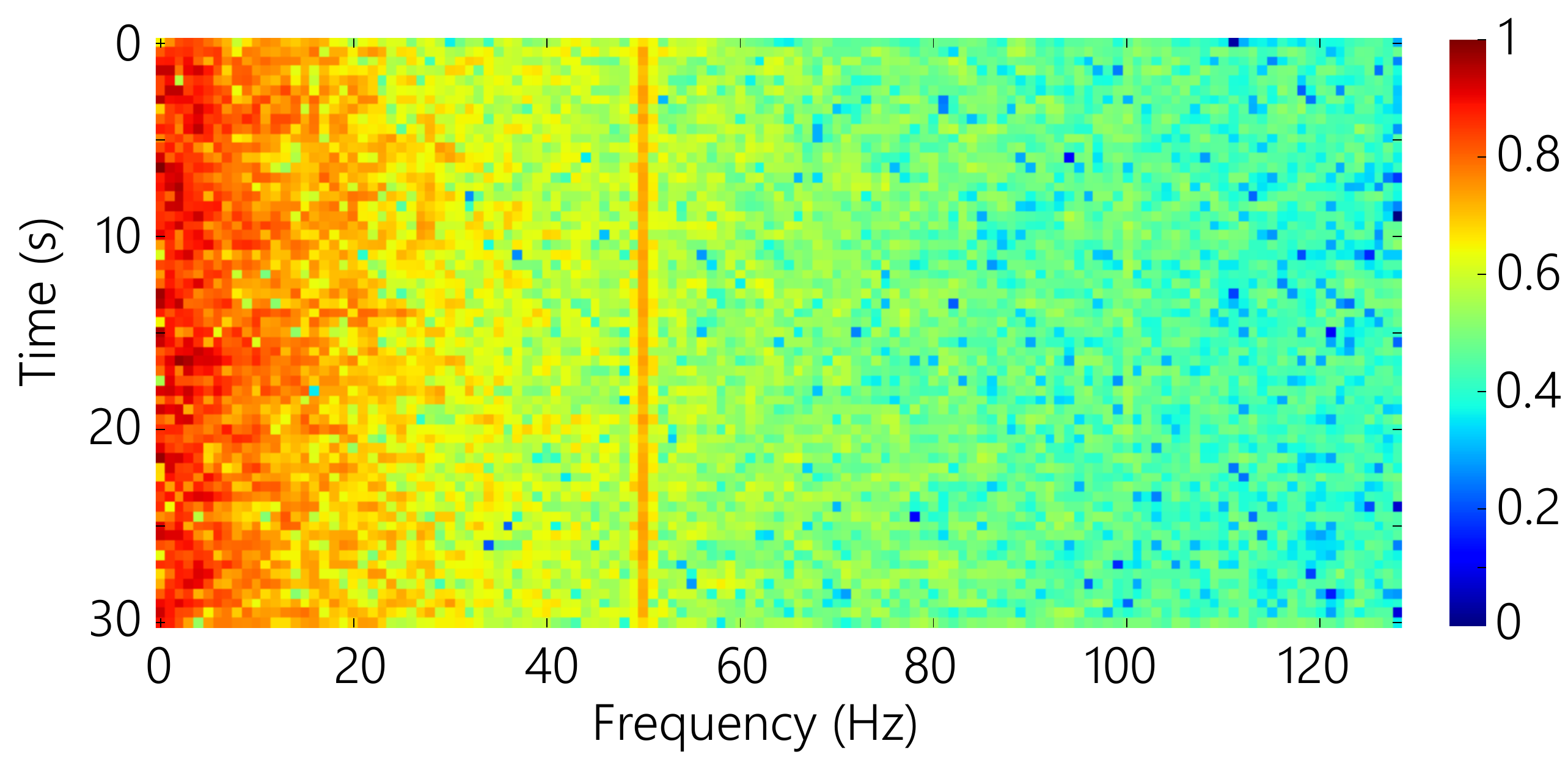}%
}

\subfloat[]{%
	\includegraphics[width=0.8\columnwidth]{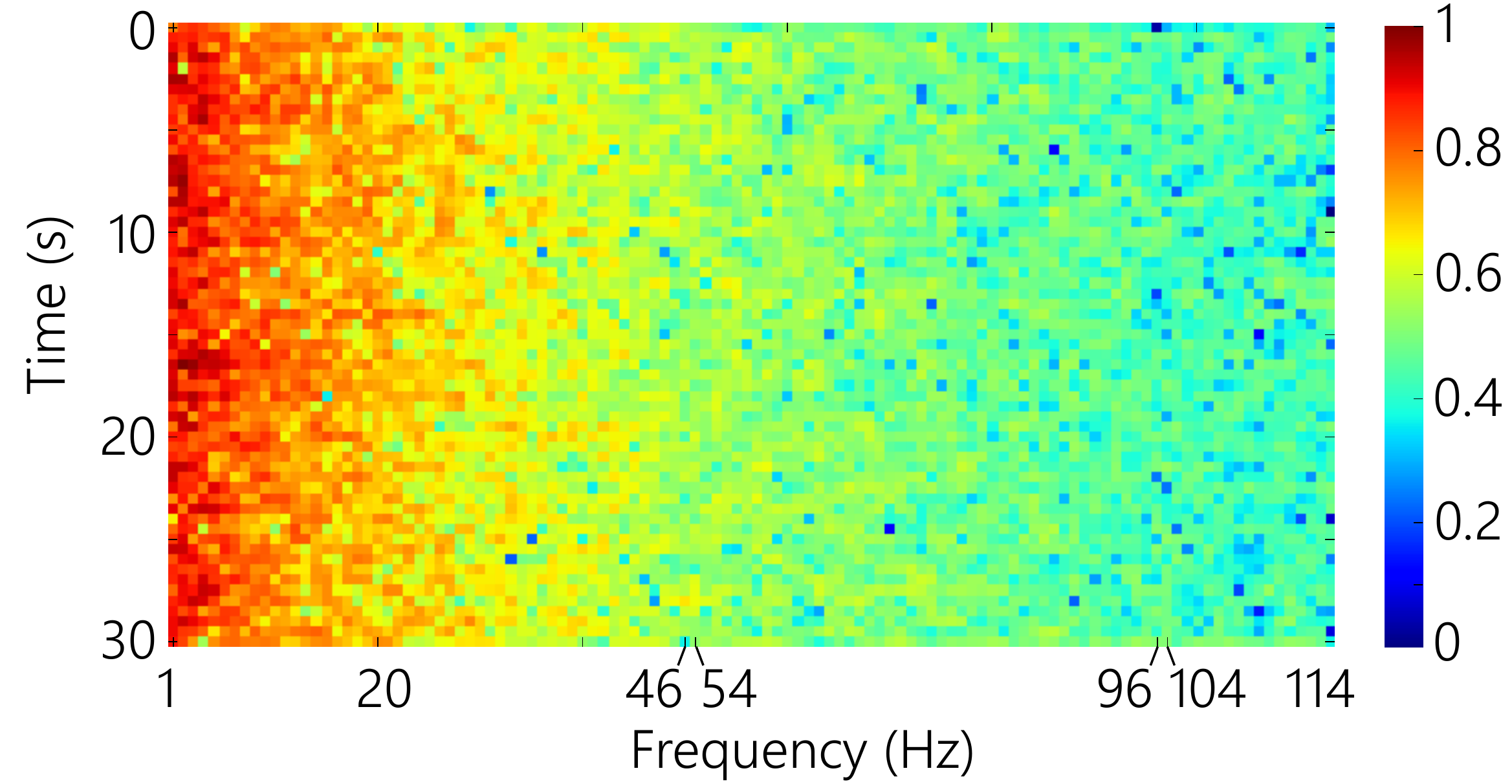}%
}

%\subfloat[]{%
%	\includegraphics[width=0.8\columnwidth]{17-norm}%
%}

\caption{Prediction performance of deep learning model.}
\label{fig:szpre:stft}
\end{figure}

Since $2$-dimensional convolutional neural network will be used in our work, it is necessary to convert raw EEG data into matrix, ie. image-like format. The conversion must be able to keep the most important information of the EEG signals. Wavelet and Fourier transform were commonly used to convert time-series EEG signals into image shape. They were also used as an effective feature extraction method for seizure detection and prediction. In this paper, we use Short-Time Fourier Transform to translate raw EEG signal into two dimensional matrix comprised of frequency and time axes. We use EEG window length of $30$~s. Most of EEG recordings were contaminated by power line noise at $50$~Hz (see Fig.~\ref{fig:szpre:stft}a) for Freiburg dataset and $60$~Hz for CHB-MIT dataset. In frequency domain, it is convenient to effectively remove the power line noise by excluding components at frequency range of $47$--$53$~Hz and $97$--$103$~Hz if power frequency is $50$~Hz and components at frequency range of $57$--$63$~Hz and $117$--$123$~Hz for power line frequency of $60$~Hz. The DC component (at $0$~Hz) was also removed. Fig.~\ref{fig:szpre:stft}b shows the STFT of a $30$-s window after removing power line noise.

%Though seizure activities commonly occur at frequencies below $30$~Hz \cite{Yuan2015,li2013szpred}, high frequency components also contain important information for seizure prediction \cite{moghim2014szpred,Sharif2017}. It is important to note that components at low frequencies have much higher magnitude than ones at high frequencies (see Fig.~\ref{fig:szpre:stft}b). This leads to the possibility that neural network approach tends to capture less information from high frequency range. To overcome this, we propose to do a standardization to bring similar average magnitude levels for all frequencies. For each patient, we calculate the average magnitude at each frequency to estimate a single scale vector for that patient. Fig.~\ref{fig:szpre:stft}c illustrates the STFT of a $30$-second EEG window after standardization.

\begin{figure}[h]
	\centering
%\hspace*{0.1in}
	\includegraphics[width=0.9\columnwidth]{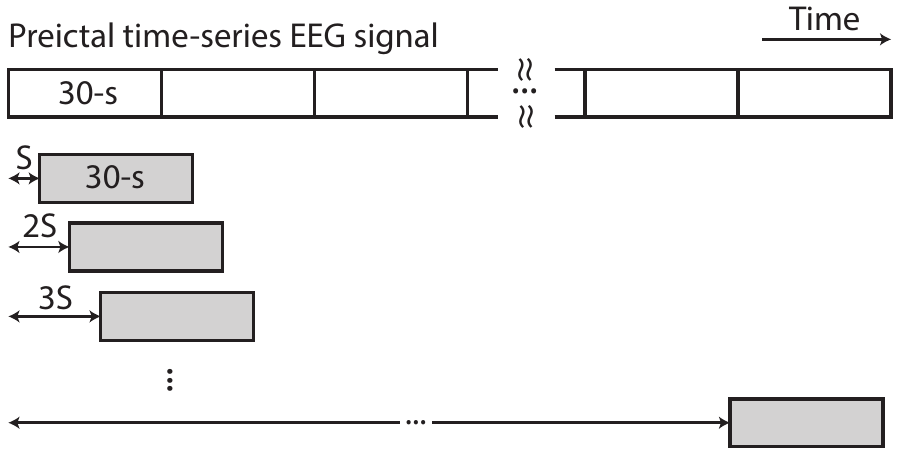}%
\caption{Generate extra preictal segments to balance the training dataset by sliding a $30$-s window along time axis at every step $S$ over preictal signals. $S$ is chosen per each subject so that there are similar number of samples per each class (preictal or interictal) in training set.}
\label{fig:szpre:uspl}
\end{figure}

One challenge to convolutional neural network is the imbalance of the dataset, i.e. much more interictal recordings than preictal ones. For example, in Freiburg dataset, we have interictal to preictal ratio per patient varies from $9.5:1$ to $15.9:1$. To overcome this, we generate more preictal segments by using overlapping technique during training phase. In particular, we create extra preictal samples for training by sliding a $30$-s window along time axis at every step $S$ over preictal time-series EEG signals~(see Fig.~\ref{fig:szpre:uspl}). $S$ is chosen per each subject so that we have similar number of samples per each class (preictal or interictal) in training set. Note that it is possible to have some extra preictal segments are the same with original ones but this would not be problematic.

\subsection{Convolutional neural network}
\begin{figure*}[h]
\centering
\includegraphics[width=0.98\textwidth]{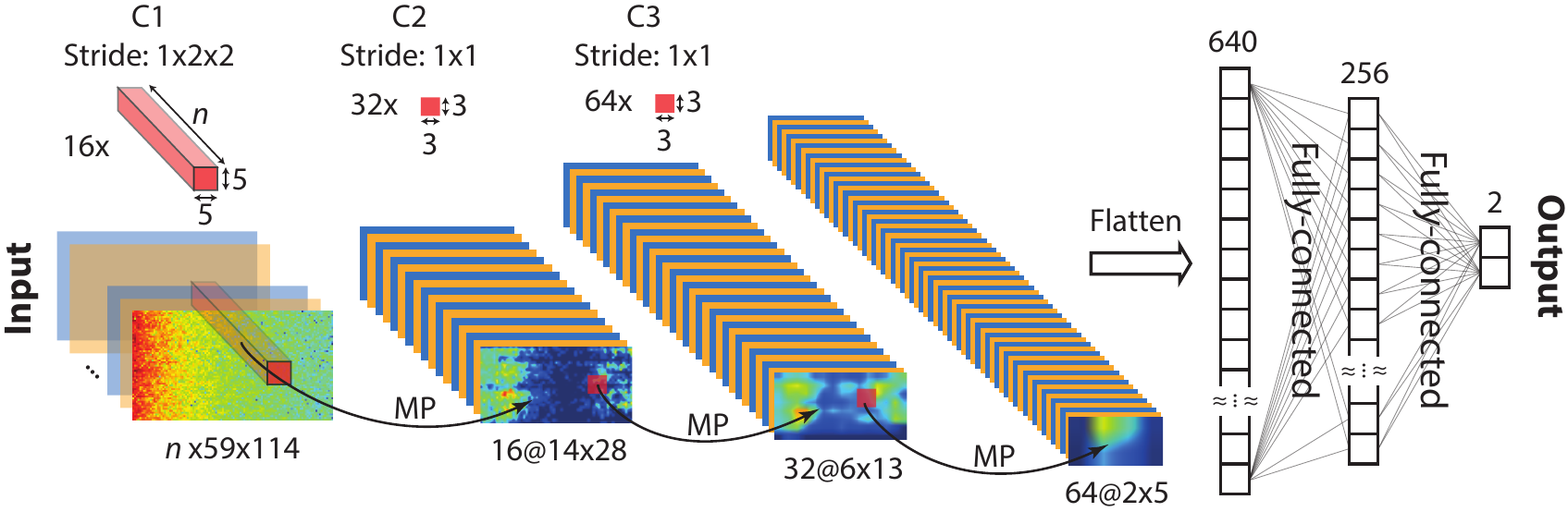}
\caption{Convolutional neural network architecture. This illustration is applied for Beefburger and CHB-MIT datasets. For Kaggle dataset, feature sizes are different due to different recording sampling rate. Input are STFT transforms of $30$s windows of raw EEG signals. There are three convolution blocks naming C1, C2 and C3. Each block consists of a batch normalization, a convolution layer with ReLU activation function, and a max pooling layer. For the sake of simplicity, max pooling layers are not shown and are noted as MP. For C1, sixteen 3-dimensional kernel with size $n \times 5 \times 5$, where $n$ is number of EEG channels, are used with stride $1 \times 2 \times 2$. ReLU activation is applied on convolution results before being sub-sampled by a max pooling layer with size $1 \times 2 \times 2$. The same steps are applied in C2 and C3 except convolution kernel size $3 \times 3$, stride $1 \times 1$ and max pooling size $2 \times 2$. Blocks C2 and C3 have $32$ and $64$ convolution kernel, respectively. Features extracted by the three convolution blocks are flatten and connected to $2$ fully-connected layers with output sizes $256$ and $2$, respectively. The former fully-connected layer uses sigmoid activation function while the latter uses soft-max activation function. Drop-out layers are placed before each of the two fully-connected layers with dropping rate of $0.5$.}
\label{fig:szpre:cnn}
\end{figure*}

\begin{figure}[h]
\centering
\includegraphics[width=0.38\textwidth]{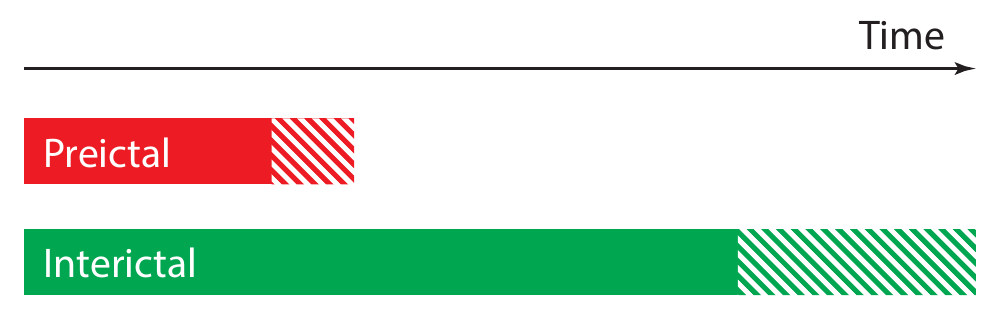}
\caption{Validation approach during training to prevent the convolutional neural network from over-fitting. $25\%$ later samples (diagonal lines) from preictal and interictal sets are used for validation and the rest for training. }
\label{fig:szpre:val}
\end{figure}

Convolutional neural networks (CNNs) have been used extensively the recent years for computer vision and natural language processing. In this paper, we use a CNN with three convolution blocks as described in Fig.~\ref{fig:szpre:cnn}. Each convolution block consists of a batch normalization, a convolution layer with a rectified linear unit (ReLU) activation function, and a max pooling layer. The batch normalization to ensure the inputs to convolution layer have zero mean and unit variance. The first convolution layer has sixteen 3-dimensional kernel with size $n \times 5 \times 5$, where $n$ is number of EEG channels, is used with stride $1 \times 2 \times 2$.  The next two convolution blocks have $32$ and $64$ convolution kernels, respectively, and both have kernel size $3 \times 3$, stride $1 \times 1$ and max pooling size $2 \times 2$. Following the three convolution blocks are two fully-connected layers with sigmoid activation and output sizes of $256$ and $2$ respectively. Drop-out layers are placed before each of the two fully-connected layers with dropping rate of $0.5$. Since the dataset for training the CNN is very limited, it is important to prevent the CNN from over-fitting. First, we keep the CNN architecture simple and shallow as described above. Second, we propose a guideline to prevent over-fitting during training the neural network. A common practice is to randomly split $20\%$ of the training set to use as a validation set. After each training epoch, a loss and/or accuracy are calculated with respects to the validation to check if the network starts to over-fit the training set. This approach works well with datasets where there is no time information involved, eg. images for classification task. For seizure prediction, it is logical to use samples from a different time period than those during training to monitor if the model starts to over-fit. In this paper, we carefully select $25\%$ later samples from preictal and interictal sets for validation and the rest for training (Fig.~\ref{fig:szpre:val}).

\subsection{Post-processing}
It is common to have isolated false positives during interictal periods. These isolated false predictions can be effectively reduced by using a discrete-time Kalman filter~\cite{parkyun2011szpred}. In this work, we employ a very simple method that is $k$-of-$n$ analysis to be consistent with the paper theme of simplicity. Particularly, for every $n$ predictions, the alarm only rises if there are at least $k$ positive predictions. As we use $30$-s windows, the CNN produces predictions every $30$~seconds. We choose $k=8$ and $n=10$ in this work.                                                                                     

\subsection{System evaluation}

\begin{figure}[h]
\centering
\includegraphics[width=1\columnwidth]{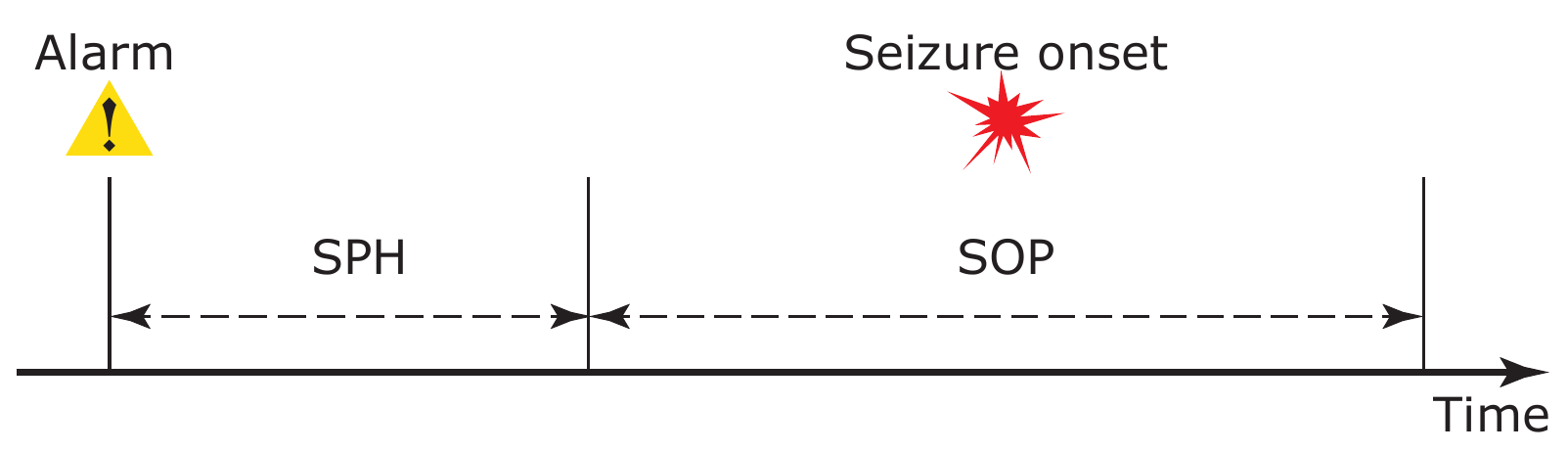}
\caption{Definition of seizure occurrence period (SOP) and seizure prediction horizon (SPH). For a correct prediction, a seizure onset must be is after the SPH and within the SOP.}
\label{fig:szpre:sph}
\end{figure}

It is non-trivial to remind how a seizure prediction system should be evaluated. Seizure prediction horizon (SPH) and seizure occurrence period (SOP) need to be defined before estimating performance metrics such as sensitivity and false prediction rate. In this paper, we follow the definition of SOP and SPH that was proposed in \cite{Maiwald2004SPH} and is illustrated in Fig.~\ref{fig:szpre:sph}. SOP is the interval where the seizure is expected to occur. The time period between the alarm and beginning of SOP is called SPH. For a correct prediction, a seizure onset must be is after the SPH and within the SOP. Likewise, a false alarm rises is when the prediction system returns a positive but there is no seizure occurring during SOP. When an alarm rises, it will last until the end of the SOP. Regarding clinical use, SPH must be long enough to allow sufficient intervention or precautions. SPH is also called intervention time \cite{BouAssi2017szpred}. In contrast, SOP should be not too long to reduce the patient's anxiety. Some works failed to mention SPH and SOP properly. In \cite{parkyun2011szpred}, the authors reported using SPH of $30$ min but based on their explanation, what they were implicitly using is SPH of $0$ min and SOP of $30$ min, ie. if a alarm occurs at any point within $30$ min before seizure onset, it is considered as a successful prediction. Similarly, authors in \cite{Zhang2016szpred} provided a different definition of SPH that is the interval between the alarm and seizure onset. Inconsistency in defining SPH and SOP make the benchmark among methods difficult and confusing.

Metrics used to test the proposed approach are sensitivity, false prediction rate under SPH of $5$~min and SOP of $30$~min. To have a robust evaluation, we follow a leave-one-out cross-validation approach for each subject. If a subject has $N$ seizures, $(N-1)$ seizures will be used for training and the withheld seizure for validation. This round is repeated $N$ times so all seizures will be used for validation exactly one time. Interictal segments are randomly split into $N$ parts. $(N-1)$ parts are used for training and the rest for validation.

\section{Results}
In this section, we are testing our approach with three datasets: (1) Freiburg iEEG dataset, (2) CHB-MIT sEEG dataset, and (3) Kaggle iEEG dataset. SOP = $30$~min and SPH = $5$~min were used in calculating all metrics in this paper. Our model is implemented in Python 2.7 using Keras 2.0 with Tensorflow 1.4.0 backend. The model was run parallelly on $4$ NVIDIA K80 graphic cards. Each fold of leave-one-out cross-validation was executed twice and average results were reported. Table~\ref{tbl:szpred:fbresult} summarizes seizure prediction results for Freiburg iEEG dataset with SOP of $30$~min and SPH of $5$~min. By applying solely the power line noise removal, prediction sensitivity is $81.4\%$, ie. $48$ out of $59$ seizures are successfully predicted. False prediction rate (FPR) is very low at $0.06$/h. Our method achieves similar sensitivity of $81.2\%$ on MIT sEEG dataset but with higher FPR of $0.16$/h~(see~Table~\ref{tbl:szpred:mitresult}). This is reasonable since scalp EEG recordings tend to be noisier than intracranial one. For Kaggle dataset, overall sensitivity is $75\%$ and FPR is $0.21$/h. It is important to note that our approach works comparably with both intracranial EEG and scalp EEG recordings without any denoising techniques except power line noise removal.

%%%%%%%%%%%%%%%%%%%%%%%%%%%%%%%%%%%%
\begin{table}[htbp]
\centering
\caption{Seizure prediction results using Freiburg iEEG dataset. SOP = $30$~min, SPH = $5$~min. $p$--value is calculated for the worst case, i.e. with minimum sensitivity and maximum false prediction rate. Our seizure prediction approach achieves significant better performance than an unspecific random predictor for all patients except Pat14.
\label{tbl:szpred:fbresult}}
\resizebox{0.49\textwidth}{!}{
	
\begin{tabular}{ l*{5}{c}  }
\toprule

	\multirow{2}{2.5em}{\centering Patient} & \multirow{2}{4em}{\centering No. of seizures} & \multirow{2}{4.5em}{\centering Interictal hours} & \multirow{2}{4.5em}{\centering$SEN$~(\%)} & \multirow{2}{4.5em}{\centering$FPR$~(/h)} & \multirow{2}{3em}{\centering$p$--value} \\ 
\\

\toprule 
\midrule

Pat1 & $4$ & $23.9$ & $100$  & $0$  & $0.000$ \\
Pat3 & $5$ & $23.9$ & $100$  & $0$  & $0.000$ \\
Pat4 & $5$ & $23.9$ & $100$  & $0$  & $0.000$ \\
Pat5 & $5$ & $23.9$ & $40$  & $0.13$  & $0.032$ \\
Pat6 & $3$ & $23.8$ & $100$  & $0$  & $0.000$ \\
Pat14 & $4$ & $22.6$ & $50$  & $0.27$  & \hl{$0.078$} \\
Pat15 & $4$ & $23.7$ & $100$  & $0.02 \pm 0.02$ & $0.000$ \\
Pat16 & $5$ & $23.9$ & $80$  & $0.17 \pm 0.13$ & $0.001$ \\
Pat17 & $5$ & $24$ & $80$  & $0$  & $0.000$ \\
Pat18 & $5$ & $24.8$ & $100$  & $0$  & $0.000$ \\
Pat19 & $4$ & $24.3$ & $50$  & $0.16$  & $0.033$ \\
Pat20 & $5$ & $24.8$ & $60$  & $0.04$  & $0.000$ \\
Pat21 & $5$ & $23.9$ & $100$  & $0$  & $0.000$ \\
\midrule
\textbf{Total} & $59$ & $311.4$ & \boldmath$81.4$ & \boldmath$0.06 \pm 0.00$ &  \\
\bottomrule
	
\end{tabular}
}
\end{table}
%%%%%%%%%%%%%%%%%%%%%%%%%%%%%%%%%%%%

%%%%%%%%%%%%%%%%%%%%%%%%%%%%%%%%%%%%
\begin{table}[htbp]
\centering
\caption{Seizure prediction results using CHB-MIT sEEG dataset. SOP = $30$~min, SPH = $5$~min. $p$--value is calculated for the worst case, i.e. with minimum sensitivity and maximum false prediction rate. Our seizure prediction approach achieves significant better performance than an unspecific random predictor for all patients except Pat9.
\label{tbl:szpred:mitresult}}
\resizebox{0.49\textwidth}{!}{
	
\begin{tabular}{ l*{5}{c}  }
\toprule

	\multirow{2}{2.5em}{\centering Patient} & \multirow{2}{4em}{\centering No. of seizures} & \multirow{2}{4.5em}{\centering Interictal hours} & \multirow{2}{4.5em}{\centering$SEN$~(\%)} & \multirow{2}{4.5em}{\centering$FPR$~(/h)} & \multirow{2}{3em}{\centering$p$--value} \\ 
\\

\toprule 
\midrule

Pat1 & $7$ & $17$ & $85.7$ & $0.235$ & $0.000$ \\
Pat2 & $3$ & $22.9$ & $33.3$ & $0$ & $0.000$ \\
Pat3 & $6$ & $21.9$ & $100$ & $0.18$ & $0.000$ \\
Pat5 & $5$ & $13$ & $80 \pm 20$ & $0.19 \pm 0.03$ & $0.010$ \\
Pat9 & $4$ & $12.3$ & $50$ & $0.12 \pm 0.12$ & \hl{$0.067$} \\
Pat10 & $6$ & $11.1$ & $33.3$ & $0.0$ & $0.025$ \\
Pat13 & $5$ & $14$ & $80$ & $0.14$ & $0.000$ \\
Pat14 & $5$ & $5$ & $80$ & $0.4$ & $0.004$ \\
Pat18 & $6$ & $23$ & $100$ & $0.28 \pm 0.02$ & $0.000$ \\
Pat19 & $3$ & $24.9$ & $100$ & $0$ & $0.000$ \\
Pat20 & $5$ & $20$ & $100$ & $0.25 \pm 0.05$ & $0.000$ \\
Pat21 & $4$ & $20.9$ & $100$ & $0.23 \pm 0.09$ & $0.000$ \\
Pat23 & $5$ & $3$ & $100$ & $0.33$ & $0.000$ \\

\midrule
\textbf{Total} & $64$ & $311.4$ & \boldmath$81.2 \pm 1.5$ & \boldmath$0.16 \pm 0.00$ &  \\
\bottomrule
	
\end{tabular}
}
\end{table}
%%%%%%%%%%%%%%%%%%%%%%%%%%%%%%%%%%%%

%%%%%%%%%%%%%%%%%%%%%%%%%%%%%%%%%%%%
\begin{table}[htbp]
\centering
\caption{Seizure prediction results using Kaggle seizure prediction competition dataset. SOP = $30$~min, SPH = $5$~min. $p$--value is calculated for the worst case, i.e. with minimum sensitivity and maximum false prediction rate. Our seizure prediction approach achieves significant better performance than an unspecific random predictor for $4$ out of $5$ dogs and for Pat1.
\label{tbl:szpred:kaggleresult}}
\resizebox{0.49\textwidth}{!}{
	
\begin{tabular}{ l*{5}{c}  }
\toprule

	\multirow{2}{2.5em}{\centering Patient} & \multirow{2}{4em}{\centering No. of seizures} & \multirow{2}{4.5em}{\centering Interictal hours} & \multirow{2}{4.5em}{\centering$SEN$~(\%)} & \multirow{2}{4.5em}{\centering$FPR$~(/h)} & \multirow{2}{3em}{\centering$p$--value} \\ 
\\

\toprule 
\midrule

Dog1 & $4$ & $80$ & $50$ & $0.19$ + $0.02$ & \hl{$0.053$} \\
Dog2 & $7$ & $83.3$ & $100$ & $0.04$ + $0.03$ & $0$ \\
Dog3 & $12$ & $240$ & $58.3$ & $0.14$ + $0.09$ & $0$ \\
Dog4 & $14$ & $134$ & $78.6$ & $0.48$ + $0.07$ & $0$ \\
Dog5 & $5$ & $75$ & $80$ & $0.08$ + $0.01$ & $0$ \\
Pat1 & $3$ & $8.3$ & $100$ & $0.42$ + $0.06$ & $0.009$ \\
Pat2 & $3$ & $7$ & $66.7$ & $0.86$ & \hl{$0.693$} \\

\midrule
\textbf{Total} & $48$ & $627.7$ & \boldmath$75$ & \boldmath$0.21  \pm 0.04$ &  \\
\bottomrule
	
\end{tabular}
}
\end{table}
%%%%%%%%%%%%%%%%%%%%%%%%%%%%%%%%%%%%

Table~\ref{tbl:szpred:compare} demonstrates a benchmark of recent seizure prediction approaches and this work. It is complicated to tell which approach is the best because each approach is usually tested with one dataset that is limited in amount of data. In other words, one approach can work well with this dataset but probably perform poorly on another dataset. Therefore, we add an extra indicator on whether same feature engineering or feature set is applied across all patients to evaluate generalization of each method. From clinical perspective, it is desirable to have long enough SPH to allow an effective therapeutic intervention and/or precautions. SOP, in the other hand, should be short to minimize the patient's anxiety \cite{Maiwald2004SPH}. Some works that implicitly used zero SPH disregarded clinical considerations, hence, could be over-estimated. Approach proposed by \cite{parkyun2011szpred} achieved a very high sensitivity of $98.3\%$ and a FPR of $0.29$/h testing with $18$ patients from Freiburg dataset. Our method yields a less sensitivity of $89.8\%$ but a better FPR of $0.17$/h. It is non-trivial to note that SPH was implicitly set to zero which means prediction at time close to or event at seizure onset can be counted as successful prediction. Likewise, researches conducted in \cite{Zhang2016szpred,Parvez2017szpred} also implied a use of zero SPH will not be compared directly to our results. Among the rest of the works listed in Table~\ref{tbl:szpred:compare}, \cite{eftekhar2014szpred} had a very good prediction sensitivity of $90.95\%$ and a low FPR of $0.06$/h under SOP = $20$~min and SPH = $10$~min. The authors in \cite{eftekhar2014szpred} were clever in fine-tuning feature extraction for each patient. This, however, leads to the need of adequate expertise and time to perform the feature engineering for new dataset. Authors in\cite{Sharif2017} applied same feature extraction technique to all patients and performed classification using SVM. This approach achieved a high sensitivity of $91.8$--$96.6\%$ and a low FPR of $0.05$--$0.08$ testing with Freiburg intracranial EEG dataset. However, there have been no works reported to successfully use similar approach on scalp EEG signals.

\begin{table*}[htbp]
	\normalsize
	
	\caption{Benchmark of recent seizure prediction approaches and this work \label{tbl:szpred:compare}}
	\resizebox{1.0\textwidth}{!}{
		\begin{tabular}{llll*{7}{c}}
			\toprule
			
			\multirow{2}{*}{Year} & \multirow{2}{*}{Authors} & \multirow{2}{*}{Dataset} & \multirow{2}{*}{Feature} & \multirow{2}{*}{Classifier} & \multirow{2}{3em}{\centering Same FE$^\triangle$} & \multirow{2}{3em}{\centering No. of\\seizures} & \multirow{2}{3em}{\centering SEN\\(\%)} & \multirow{2}{3em}{\centering FPR\\(/h)} & \multirow{2}{*}{SOP} & \multirow{2}{*}{SPH} \\ \\
			\toprule  \midrule

2004 & \multirow{2}{6em}{Maiwald\\ \textit{et al}~\cite{Maiwald2004SPH}} & FB$^\dagger$, $21$ pat. & \multirow{2}{11em}{Dynamical\\similarity index} & \multirow{2}{6em}{\centering Threshold crossing} & Yes & $88$ & $42$ & $<0.15$ & $30$~min & $2$~min \\ \\ [1.5ex]
2006 & \multirow{2}{6em}{Winterhalder\\ \textit{et al}~\cite{winterhalder2006szpred}} & FB$^\dagger$, $21$ pat. & \multirow{2}{11em}{Phase coherence,\\lag synchronization} & \multirow{2}{6em}{\centering Threshold crossing} & No & $88$ & $60$ & $0.15$ & $30$~min & $10$~min \\ \\ [1.5ex]
2011 & \multirow{2}{6em}{Park\\ \textit{et al}~\cite{parkyun2011szpred}} & FB$^\dagger$, $18$ pat. & \multirow{2}{11em}{Univariate\\spectral power} & SVM & Yes & $80$ & $98.3$ & $0.29$ & $30$~min & $0$$^\ast$ \\ \\ [1.5ex]
2012 & \multirow{2}{6em}{Gadhoumi\\ \textit{et al}~\cite{Gadhoumi2012szpred}} & MNI$^\ddagger$, $6$ pat. & \multirow{2}{11em}{Wavelet energy, entropy} & \multirow{2}{6em}{\centering Discriminant analysis} & No & $38$ & $88.9$ & $0.30$ & N/A & $22$~min \\ \\ [1.5ex]
2013 & \multirow{2}{6em}{Li\\ \textit{et al}~\cite{li2013szpred}} & FB$^\dagger$, $21$ pat. & Spike rate & \multirow{2}{6em}{\centering Threshold crossing} & Yes & $87$ & $72.7$ & $0.11$ & $50$~min & $10$~s \\ \\ [1.5ex]
2014 & \multirow{2}{6em}{Zheng\\ \textit{et al}~\cite{zheng2014szpred}} & FB$^\dagger$, $10$ pat. & \multirow{2}{11em}{Mean phase coherence} & \multirow{2}{6em}{\centering Threshold crossing} & No & $50$ & $>70$ & $<0.15$ & $30$~min & $10$~min \\ \\ [1.5ex]
2014 & \multirow{2}{6em}{Eftekhar\\ \textit{et al}~\cite{eftekhar2014szpred}} & FB$^\dagger$, $21$ pat. & \multirow{2}{11em}{Multiresolution N-gram} & \multirow{2}{6em}{\centering Threshold crossing} & No & $87$ & $90.95$ & $0.06$ & $20$~min & $10$~min \\ \\ [1.5ex]
2014 & \multirow{2}{6em}{Aarabi\\ \textit{et al}~\cite{aarabi2014szpred}} & FB$^\dagger$, $21$ pat. & \multirow{2}{11em}{Bayesian inversion of power spectral density} & \multirow{2}{6em}{\centering Rule-based decision} & Yes & $87$ & $87.07$ & $0.20$ & $30$~min & $10$~s \\ \\ [1.5ex]
2016 & \multirow{2}{6em}{Zhang\\ \textit{et al}~\cite{Zhang2016szpred}} & FB$^\dagger$, $18$ pat. & \multirow{2}{11em}{Power spectral density ratio} & SVM & No & $80$ & $100$ & $0.03$ & $50$~min & $0$$^\ast$ \\ \\ [1.5ex]
2016 & \multirow{2}{6em}{Zhang\\ \textit{et al}~\cite{Zhang2016szpred}} & MIT$^\diamondsuit$, $17$ pat. & \multirow{2}{11em}{Power spectral density ratio} & SVM & No & $76$ & $98.68$ & $0.05$ & $50$~min & $0$$^\ast$ \\ \\ [1.5ex]
2017 & \multirow{2}{6em}{Parvez\\ \textit{et al}~\cite{Parvez2017szpred}} & FB$^\dagger$, $21$ pat. & \multirow{2}{11em}{Phase-match error, deviation, fluctuation} & LS-SVM & Yes & $87$ & $95.4$ & $0.36$ & $30$~min & $0$$^\ast$ \\ \\ [1.5ex]
2017 & \multirow{2}{6em}{Sharif\\ \textit{et al}~\cite{Sharif2017}} & FB$^\dagger$, $19$ pat. & \multirow{2}{11em}{Fuzzy rules on Poincare plane} & SVM & Yes & $83$ & $91.8$--$96.6$ & $0.05$--$0.08$ & $15$min & $2$--$42$~min \\ \\ [1.5ex]
2017 & \multirow{2}{6em}{Aarabi\\ \textit{et al}~\cite{Aarabi2017szpred}} & FB$^\dagger$, $10$ pat. & \multirow{2}{11em}{Univariate and bivariate features} & \multirow{2}{6em}{\centering Rule-based decision} & Yes & $28$ & $86.7$ & $0.126$ & $30$~min & $10$~s \\ \\ [1.5ex]
2017 & This work & FB$^\dagger$, $13$ pat. & \multirow{2}{11em}{Short-Time Fourier~Transform} & \multirow{2}{6em}{\centering CNN} & Yes & $59$ & $81.4$ & $0.06$ & $30$~min & $5$~min \\ \\ [1.5ex]
2017 & This work & MIT$^\diamondsuit$, $13$ pat. & \multirow{2}{11em}{Short-Time Fourier~Transform} & \multirow{2}{6em}{\centering CNN} & Yes & $64$ & $81.2$ & $0.16$ & $30$~min & $5$~min \\ \\ [1.5ex]
2017 & This work & \multirow{2}{8em}{Kaggle$^\nabla$,\\$5$ dogs, $2$ pat.} & \multirow{2}{11em}{Short-Time Fourier~Transform} & \multirow{2}{6em}{\centering CNN} & Yes & $48$ & $75$ & $0.21$ & $30$~min & $5$~min \\ \\

			\bottomrule 
			
	\end{tabular}}{\scriptsize
		\begin{tablenotes}
			\item[] {$^\triangle$~Same feature engineering across all patients. "No" means feature engineering is carefully tailored for each patient.}
            \item[] {$^\ast$~The authors implicitly used zero SPH, disregarded clinical considerations, hence, the results could be over-estimated.}
			\item[] {$^\dagger$~Freiburg Hospital intracranial EEG (iEEG) dataset.}
			\item[] {$^\ddagger$~Montreal Neurological Institute intracranial EEG (iEEG) dataset.}
            \item[] {$^\diamondsuit$~Massachusetts Institute of Technology scalp EEG (sEEG) dataset.}         
            \item[] {$^\nabla$~Kaggle American Epilepsy Society Seizure Prediction Challenge dataset.}      
            
	\end{tablenotes}}
\end{table*}

\section{Discussion}

Information extracted from EEG signals in frequency and time (synchronization) domains has been used widely to predict seizures. As seen in Table~\ref{tbl:szpred:compare}, sensitivity and false prediction have improved over time. This paper proposed a novel way to exploit both frequency and time aspects of EEG signals without handcraft feature engineering. Short-Time Fourier Transform of an EEG window has two axes of frequency and time. A $2$--dimensional convolution filter was slid throughout the STFT to collect the changes in both frequency and time of EEG signals. This is where the beauty of convolutional neural network comes in. The filter weights are automatically adjusted during the training phase and the CNN acts like a feature extraction method in a automatic fashion.

We also compare the prediction performance of our approach with an unspecific random predictor. Given an FPR, the probability to raise an alarm in an SOP can be approximated by~\cite{schelter2006testing}
\begin{equation}
P \approx 1-e^{-\textrm{FPR} \cdot \textrm{SOP}}
\end{equation}

Therefore, probability of predicting at least $k$ of $K$ independent seizures by chance is given by
\begin{equation}
p = \sum_{j \ge k} {K\choose j} P^{j}(1-P)^{K-j}
\end{equation}

We calculated $p$ value for each patient by using FPR of that patient and the number of predicted seizures ($k$) by our method. If $p$ is less than $0.05$, we can conclude that our prediction method is significantly better than a random predictor at significant level of $0.05$. Tables~\ref{tbl:szpred:fbresult}~and~\ref{tbl:szpred:mitresult} have shown that our prediction method achieve significantly superior to an unspecific random predictor for all patients except Pat14 in Freiburg dataset and Pat9 in CHB-MIT dataset. It is worth reminding that Freiburg dataset is intracranial EEG while CHB-MIT dataset is scalp EEG. In other words, our method works well with both types of EEG signals. Regarding Kaggle dataset, our method results in significantly better performance compared to a random predictor for $4$ out of $5$ canines~(see Table~\ref{tbl:szpred:kaggleresult}) and for Pat1. 

%This work further improves the capability in feature extraction of CNN with regards to seizure prediction by standardizing the amplitudes of STFT components across the whole frequency range. Though seizure activities commonly occur at frequencies below $30$~Hz \cite{Yuan2015,li2013szpred}, high frequency components also contain important information for seizure prediction \cite{moghim2014szpred,Sharif2017}. Amplitude standardization prevent high frequency components from being dominated by those in lower frequency ranges. The two datasets tested in this paper have sampling rate of $256$~Hz that means highest frequency after STFT is just $128$~Hz. We suggest our approach would give better outcomes with higher sampling rate.

As seizure characteristics may change over time, calibration of seizure prediction algorithm is necessary. Minimum feature engineering brings great advantage that it does not require an expert to carefully extract and select optimum features for the prediction task. Hence, it allows faster and more frequent updates so that patients are able to benefit the most from the seizure prediction algorithm. Also, minimum feature engineering also allows the seizure prediction available to more patients. Since feature extraction task is taken by CNN, neuro-physiologists and clinical staff can spend more time in monitoring and recording EEG signals for diagnostic purpose and/or training data collection.

\begin{figure}[h]
\centering
\includegraphics[width=0.92\columnwidth]{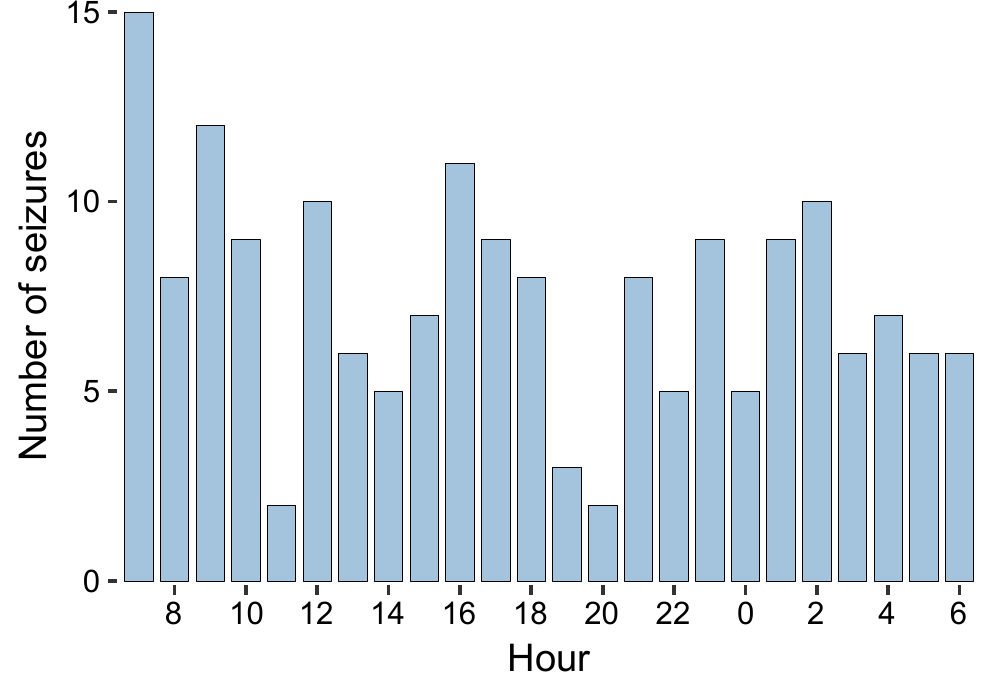}
\caption{Distribution of number of seizures over time of day across patients for CHB-MIT dataset. Most seizures occur in the early morning. Two lower peaks are around $4$~p.m. and $2$~a.m.}
\label{fig:szpre:szdist}
\end{figure}

Our method can be further improved by non-EEG data such as information of time when seizures occurs. Epileptic seizures have been shown to have biases in distribution over time at various intervals that can be as long as year or as short as hour~\cite{Griffiths1938rhythm}. Importantly, the authors in \cite{Griffiths1938rhythm} have shown that there are more incidences of seizure around sunrise, noon and midnight in their dataset of $101$ patients with $39,929$ seizures. However, this pattern is patient-specific and can be very different from patient to patient. Adopting the same observation, authors in \cite{Karoly2017circadian} leverage this pattern to significantly improve their seizure forecasting system. Unfortunately, the three datasets investigated in this paper is not long enough to assess if time of day information is useful because maximum recording period per patient is $3$ days. Nevertheless, it is still worth to see how incidences of seizure are distributed over time of day across patients in the CHB-MIT dataset, the only dataset that we can access time of seizure occurrence. Fig.~\ref{fig:szpre:szdist} shows greatest incidence in the early morning, and two lower peaks around $4$~p.m. and $2$~a.m. 

\section{Conclusion}

Seizure prediction capability has been studied and improved over the last four decades. A perfect prediction is yet available but with current prediction performance, it is useful to provide the patients with warning message so they can take some precautions for their safety. This paper proposed a novel approach of using convolutional neural network with minimum feature engineering. The proposed shown its good generalization when working well with both intracranial EEG and scalp EEG. This brings opportunity to more patients to possess a seizure prediction device that can help them have a more manageable life.

% or


%\appendices

%\section{}

\section{Acknowledgment}

The authors appreciate Dr Benjamin H. Brinkmann support from Mayo Systems Electrophysiology Lab for providing information on some unlabeled datasets. The authors also thank Dr Farzaneh Shayegh for sharing her thoughts on Freiburg Hospital dataset. N. Truong acknowledges The Commonwealth Scientific and Industrial Research Organisation (CSIRO) partial financial support via a PhD Scholarship, PN~50041400. J. Yang acknowledges National Natural Science Foundation of China for their financial support under Grant 61501332.

\ifCLASSOPTIONcaptionsoff
  \newpage
\fi



%\bibliographystyle{IEEEtran}
%\cleardoublepage
\bibliographystyle{IEEEtran}
\bibliography{MyCollection}

\begin{thebibliography}{10}
\providecommand{\url}[1]{#1}
\csname url@samestyle\endcsname
\providecommand{\newblock}{\relax}
\providecommand{\bibinfo}[2]{#2}
\providecommand{\BIBentrySTDinterwordspacing}{\spaceskip=0pt\relax}
\providecommand{\BIBentryALTinterwordstretchfactor}{4}
\providecommand{\BIBentryALTinterwordspacing}{\spaceskip=\fontdimen2\font plus
\BIBentryALTinterwordstretchfactor\fontdimen3\font minus
  \fontdimen4\font\relax}
\providecommand{\BIBforeignlanguage}[2]{{%
\expandafter\ifx\csname l@#1\endcsname\relax
\typeout{** WARNING: IEEEtran.bst: No hyphenation pattern has been}%
\typeout{** loaded for the language `#1'. Using the pattern for}%
\typeout{** the default language instead.}%
\else
\language=\csname l@#1\endcsname
\fi
#2}}
\providecommand{\BIBdecl}{\relax}
\BIBdecl

\bibitem{Ramgopal2014szloop}
S.~Ramgopal, S.~Thome-Souza, M.~Jackson, N.~E. Kadish, I.~{S{\'{a}}nchez
  Fern{\'{a}}ndez}, J.~Klehm, W.~Bosl, C.~Reinsberger, S.~Schachter, and
  T.~Loddenkemper, ``{Seizure detection, seizure prediction, and closed-loop
  warning systems in epilepsy},'' \emph{Epilepsy {\&} Behavior}, vol.~37, pp.
  291--307, aug 2014.

\bibitem{Gadhoumi2016}
K.~Gadhoumi, J.-M. Lina, F.~Mormann, and J.~Gotman, ``{Seizure prediction for
  therapeutic devices: A review},'' \emph{Journal of Neuroscience Methods},
  vol. 260, pp. 270--282, 2016.

\bibitem{BouAssi2017szpred}
E.~{Bou Assi}, D.~K. Nguyen, S.~Rihana, and M.~Sawan, ``{Towards accurate
  prediction of epileptic seizures: A review},'' \emph{Biomedical Signal
  Processing and Control}, vol.~34, pp. 144--157, 2017.

\bibitem{rogowski1981szpred}
Z.~Rogowski, I.~Gath, and E.~Bental, ``{On the prediction of epileptic
  seizures},'' \emph{Biological Cybernetics}, vol.~42, no.~1, pp. 9--15, 1981.

\bibitem{salant1998szpred}
Y.~Salant, I.~Gath, and O.~Henriksen, ``{Prediction of epileptic seizures from
  two-channel EEG},'' \emph{Medical and Biological Engineering and Computing},
  vol.~36, no.~5, pp. 549--556, 1998.

\bibitem{Maiwald2004SPH}
T.~Maiwald, M.~Winterhalder, R.~Aschenbrenner-Scheibe, H.~U. Voss,
  A.~Schulze-Bonhage, and J.~Timmer, ``{Comparison of three nonlinear seizure
  prediction methods by means of the seizure prediction characteristic},''
  \emph{Physica D: Nonlinear Phenomena}, vol. 194, no. 3-4, pp. 357--368, 2004.

\bibitem{winterhalder2006szpred}
M.~Winterhalder, B.~Schelter, T.~Maiwald, A.~Brandt, A.~Schad,
  A.~Schulze-Bonhage, and J.~Timmer, ``{Spatio-temporal patient--individual
  assessment of synchronization changes for epileptic seizure prediction},''
  \emph{Clinical Neurophysiology}, vol. 117, no.~11, pp. 2399--2413, 2006.

\bibitem{zheng2014szpred}
Y.~Zheng, G.~Wang, K.~Li, G.~Bao, and J.~Wang, ``{Epileptic seizure prediction
  using phase synchronization based on bivariate empirical mode
  decomposition},'' \emph{Clinical Neurophysiology}, vol. 125, no.~6, pp.
  1104--1111, 2014.

\bibitem{Parvez2017szpred}
M.~Z. Parvez and M.~Paul, ``{Seizure prediction using undulated global and
  local features},'' \emph{IEEE Transactions on Biomedical Engineering},
  vol.~64, no.~1, pp. 208--217, jan 2017.

\bibitem{parkyun2011szpred}
Y.~Park, L.~Luo, K.~K. Parhi, and T.~Netoff, ``{Seizure prediction with
  spectral power of EEG using cost-sensitive support vector machines},''
  \emph{Epilepsia}, vol.~52, no.~10, pp. 1761--1770, 2011.

\bibitem{EEGFB}
\BIBentryALTinterwordspacing
``{EEG Database at the Epilepsy Center of the University Hospital of Freiburg,
  Germany}.'' [Online]. Available:
  \url{https://epilepsy.uni-freiburg.de/freiburg-seizure-prediction-project/eeg-database/}
\BIBentrySTDinterwordspacing

\bibitem{Zhang2016szpred}
Z.~Zhang and K.~K. Parhi, ``{Low-complexity seizure prediction from iEEG/sEEG
  using spectral power and ratios of spectral power},'' \emph{IEEE Transactions
  on Biomedical Circuits and Systems}, vol.~10, no.~3, pp. 693--706, 2016.

\bibitem{aarabi2014szpred}
A.~Aarabi and B.~He, ``{Seizure prediction in hippocampal and neocortical
  epilepsy using a model-based approach},'' \emph{Clinical Neurophysiology},
  vol. 125, no.~5, pp. 930--940, 2014.

\bibitem{Aarabi2017szpred}
------, ``{Seizure prediction in patients with focal hippocampal epilepsy},''
  \emph{Clinical Neurophysiology}, 2017.

\bibitem{eftekhar2014szpred}
A.~Eftekhar, W.~Juffali, J.~El-Imad, T.~G. Constandinou, and C.~Toumazou,
  ``{Ngram-derived pattern recognition for the detection and prediction of
  epileptic seizures},'' \emph{PloS one}, vol.~9, no.~6, p. e96235, 2014.

\bibitem{Sharif2017}
B.~Sharif and A.~H. Jafari, ``{Prediction of epileptic seizures from EEG using
  analysis of ictal rules on Poincar{\'{e}} plane},'' \emph{Computer Methods
  and Programs in Biomedicine}, vol. 145, pp. 11--22, 2017.

\bibitem{Gadhoumi2012szpred}
K.~Gadhoumi, J.-M. Lina, and J.~Gotman, ``{Discriminating preictal and
  interictal states in patients with temporal lobe epilepsy using wavelet
  analysis of intracerebral EEG},'' \emph{Clinical Neurophysiology}, vol. 123,
  no.~10, pp. 1906--1916, oct 2012.

\bibitem{li2013szpred}
S.~Li, W.~Zhou, Q.~Yuan, and Y.~Liu, ``{Seizure prediction using spike rate of
  intracranial EEG},'' \emph{IEEE Transactions on Neural Systems and
  Rehabilitation Engineering}, vol.~21, no.~6, pp. 880--886, 2013.

\bibitem{mirowski2008comparing}
P.~W. Mirowski, Y.~LeCun, D.~Madhavan, and R.~Kuzniecky, ``{Comparing SVM and
  convolutional networks for epileptic seizure prediction from intracranial
  EEG},'' \emph{2008 IEEE Workshop on Machine Learning for Signal Processing},
  pp. 244--249, 2008.

\bibitem{shoeb2009application}
A.~H. Shoeb, ``{Application of machine learning to epileptic seizure onset
  detection and treatment},'' Ph.D. dissertation, Massachusetts Institute of
  Technology, 2009.

\bibitem{BenjaminKaggleSzPred2014}
B.~H. Brinkmann, J.~Wagenaar, D.~Abbot, P.~Adkins, S.~C. Bosshard, M.~Chen,
  Q.~M. Tieng, J.~He, F.~J. Mu{\~{n}}oz-Almaraz, P.~Botella-Rocamora, J.~Pardo,
  F.~Zamora-Martinez, M.~Hills, W.~Wu, I.~Korshunova, W.~Cukierski, C.~Vite,
  E.~E. Patterson, B.~Litt, and G.~A. Worrell, ``{Crowdsourcing reproducible
  seizure forecasting in human and canine epilepsy},'' \emph{Brain}, vol. 139,
  no.~6, pp. 1713--1722, 2016.

\bibitem{schelter2006testing}
B.~Schelter, M.~Winterhalder, T.~Maiwald, A.~Brandt, A.~Schad,
  A.~Schulze-Bonhage, and J.~Timmer, ``{Testing statistical significance of
  multivariate time series analysis techniques for epileptic seizure
  prediction},'' \emph{Chaos: An Interdisciplinary Journal of Nonlinear
  Science}, vol.~16, no.~1, p. 13108, 2006.

\bibitem{Griffiths1938rhythm}
G.~M. Griffiths and J.~T. Fox, ``{Rhythm in epilepsy},'' \emph{The Lancet},
  vol. 232, no. 5999, pp. 409--416, 1938.

\bibitem{Karoly2017circadian}
P.~J. Karoly, H.~Ung, D.~B. Grayden, L.~Kuhlmann, K.~Leyde, M.~J. Cook, and
  D.~R. Freestone, ``{The circadian profile of epilepsy improves seizure
  forecasting},'' \emph{Brain}, vol. 140, no.~8, pp. 2169--2182, 2017.

\end{thebibliography}
%\bibliography{Mendeley}


%\begin{IEEEbiography}{Nhan Truong}
%\end{IEEEbiography}

%\begin{IEEEbiographynophoto}{Nhan Truong}

%\newpage

%\begin{IEEEbiography}{Omid Kavehei}
%\end{IEEEbiography}


%\vfill


\end{document}